\documentclass{article}

\usepackage[dblblindworkshop,final]{neurips_2026}
\workshoptitle{SLMs for Agentic Systems, Paris, France, 2026}

\usepackage[utf8]{inputenc}
\usepackage[T1]{fontenc}
\usepackage[hidelinks]{hyperref}
\usepackage{url}
\usepackage{booktabs}
\usepackage{amsmath}
\usepackage{graphicx}
\usepackage{xcolor}
\usepackage[section,below]{placeins}
\usepackage{pgfplots}
\pgfplotsset{compat=1.18}

\newcommand{\qet}{ET}
\newcommand{\qit}{IT}
\newcommand{\qep}{EP}

\title{Quantization Effects on Tool-Failure Recovery Vary Across Prompts and Evaluation Designs}

\author{%
  Yuhe Hu\\
  Duke University\\
  \texttt{yh353@duke.edu}\\
}

\begin{document}
\raggedbottom
\maketitle

\begin{abstract}
Post-training quantization reduces the cost of deploying language-model
agents, but its effect on recovery from temporary tool failures can
depend on how recovery is evaluated. We compare 8-bit and 4-bit variants
of Llama-3.1-8B-Instruct and Qwen2.5-7B-Instruct on twenty deterministic
tool-use tasks and five prompts. The 8-bit--4-bit recovery comparison
changes direction across prompts and evaluation targets. On tasks that
both variants complete without faults under the same prompt, the
difference ranges from $0$ to $+20.2$ percentage points for Llama and
from $-50.0$ to $+35.0$ points for Qwen. Full-pipeline point estimates
favor 8-bit Llama under all five prompts, whereas the Qwen comparison
changes direction across prompts. The evaluation target can also reverse
the result. For Llama under one prompt, scoring each variant only on its
own clean-passing tasks favors 4-bit by $17.5$ points; scoring the same
tasks for both variants gives no difference, while scoring the full
pipeline favors 8-bit by $28.3$ points. Executor leniency is a third
such choice. Rescoring the same logs with strict output parsing, which
8-bit Llama violates far more often than 4-bit Llama under that prompt,
turns that $+28.3$ into $-15.0$ while leaving Qwen essentially
unchanged. These findings show
that one prompt, one screened task set, and one scoring policy do not
establish a stable conclusion about quantized-agent robustness.
Evaluations should compare variants on matched tasks, report
full-pipeline success for deployment decisions, state the scoring
policy, and quantify uncertainty across tasks rather than injected fault
sites.
\end{abstract}

\section{Introduction}

\begin{figure}[t]
\centering
\includegraphics[width=\linewidth]{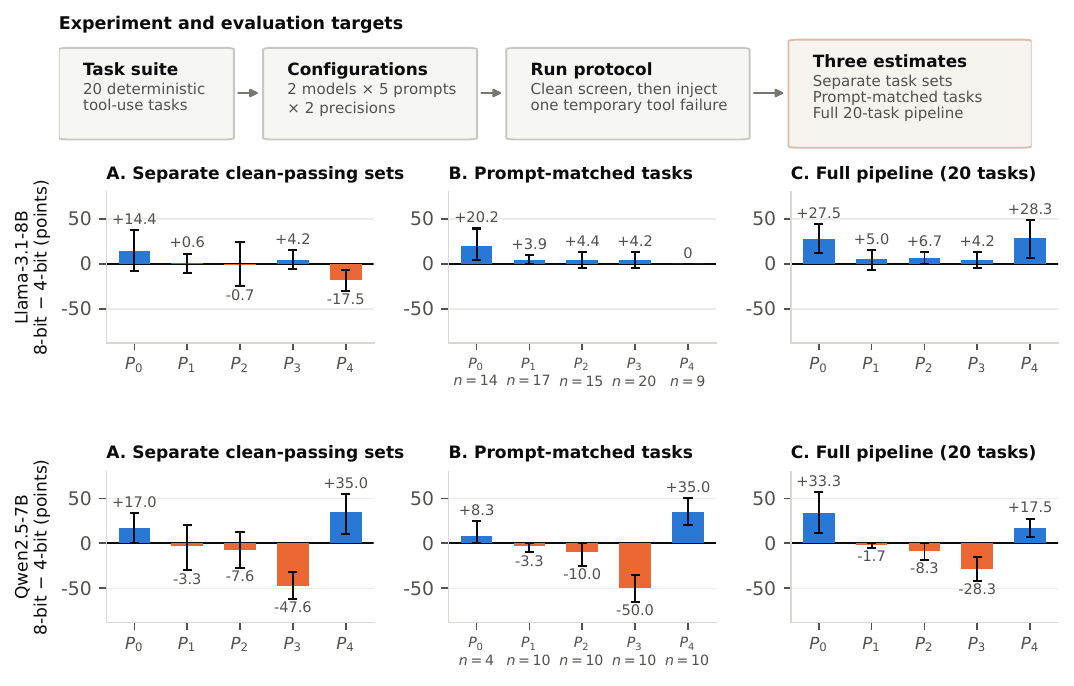}
\caption{\textbf{The measured 8-bit--4-bit difference varies across
models, prompts, and evaluation targets.} \textbf{A.} Each variant's own
clean-passing tasks. \textbf{B.} Tasks that both variants pass cleanly
under the same prompt; the number of matched tasks is shown below each
prompt. \textbf{C.} All twenty tasks, with a clean failure scored as
zero. Error bars are 95\% task-bootstrap intervals. Panel A uses a
two-sample bootstrap because the task sets can differ; Panels B and C
use paired bootstraps.}
\label{fig:collapse}
\end{figure}

Post-training quantization reduces the memory and compute required to run
language-model agents, making 4-bit checkpoints attractive when inference
cost matters \citep{frantar2022gptq,dettmers2022case,llamacpp}. Existing
evaluations of compressed agents primarily report task completion.
Deployment also requires the agent to respond appropriately when
a tool is temporarily unavailable. It should retry when recovery is
possible, avoid inventing a result, and complete the task after service
returns. This paper asks how 8-bit and 4-bit quantization affect that
recovery behavior.

Two sources of variation make this question harder than a single success
rate suggests. First, recovery behavior can change with the instructions
that specify when to retry or stop. Second, fault-recovery experiments
often inject failures only into tasks that the tested configuration first
completes without faults. If that clean screen is recomputed for each
model, precision, and prompt, the reported rates may describe different
tasks. A comparison can then combine a recovery difference with a change
in which tasks were selected.

We evaluate Llama-3.1-8B-Instruct and Qwen2.5-7B-Instruct at 8 and 4 bits
on twenty deterministic order-management tasks, five prompts, and
transient failures injected at every valid site on the required tool
path. We analyze each run in three ways (Figure~\ref{fig:collapse}):
recovery on each configuration's own clean-passing tasks; recovery on
tasks passed by both precisions under the same prompt; and full-pipeline
success over all twenty tasks, with clean failures scored as failures.
Llama is also evaluated at 3 bits, but the prompt interaction at that
precision is exploratory and appears in the appendix.

The main result is that neither 8-bit nor 4-bit quantization provides a
consistent recovery advantage across models and prompts. On
prompt-matched tasks, Llama's 8-bit--4-bit estimates range from $0$ to
$+20.2$ points, whereas Qwen's range from $-50.0$ under $P_3$ to $+35.0$
under $P_4$. Full-pipeline estimates favor 8-bit Llama under all five
prompts but change sign across Qwen prompts. These results do not support
a model-independent claim that one of the two precisions is more robust
to temporary tool failures.

The analysis also shows why the evaluation target must be stated
explicitly. For Llama/$P_4$, scoring each precision on its own
clean-passing tasks favors 4-bit by $17.5$ points. The difference is zero
on the nine tasks passed by both precisions under $P_4$, while
full-pipeline success favors 8-bit by $28.3$ points because the 8-bit
model passes many more tasks cleanly. In contrast, Qwen/$P_3$ and
Qwen/$P_4$ retain large, oppositely directed differences after matching
the tasks, so task selection cannot explain all observed variation.
The scoring policy of the harness has the same power. Our executor
accepts arithmetic expressions submitted in place of a literal amount,
under $P_4$ 8-bit Llama relies on this tolerance far more than 4-bit
Llama does, and rescoring the logs under a strict policy reverses the
Llama/$P_4$ comparison on every target (Section~\ref{sec:scoring}).

Our contribution is therefore empirical and methodological. We connect
the study of compressed agents with controlled tool-failure evaluation,
show that the measured quantization effect varies across two models,
five prompts, and two scoring policies, and identify when separate
clean screening or executor leniency changes the interpretation. The
resulting recommendation is to report matched-task recovery for
conditional comparisons and full-pipeline success for deployment
decisions, to state the scoring policy, and to use tasks as the unit of
uncertainty.

\section{Related work}

Work on compressed agents and work on tool-failure recovery address
different parts of our question. ACBench \citep{dong2025acbench} evaluates
agentic capabilities under compression, and \citet{jang2026flatscore}
show that aggregate scores can hide quantization-induced failures in
$\tau^2$-bench \citep{barres2025tau2,yao2024taubench}. These studies do
not isolate recovery from injected temporary tool failures. ToolMaze
\citep{zhu2026toolmaze} provides the fault taxonomy and recovery framing
used here, but evaluates full-precision agents under fixed prompts. Our
study connects these lines by comparing quantized variants under
controlled tool failures and multiple prompt formulations. ACBench and
ToolMaze each report a single success rate on tasks screened per
configuration. We show on released logs that the clean screen, the
evaluation target, and the executor's parsing policy can each change
the sign of a quantization comparison, and we provide matched-task and
full-pipeline estimators that keep such comparisons interpretable.

Prompt sensitivity is well documented. Brittlebench
\citep{romanou2026brittlebench} measures ranking changes under
semantics-preserving perturbations, while \citet{hua2025flawartifact}
study evaluation artifacts in measured prompt sensitivity. Our prompts
are not controlled paraphrases, so we interpret prompt differences as
variation across deployment instructions rather than linguistic
robustness. Recovery harnesses such as DARC \citep{wang2026darc} would
need the same matched-task analysis when used to compare model variants.

\section{Evaluation targets}
\label{sec:estimands}

Let $S(a,p)$ be the set of tasks that pass a clean screen at precision $a$
under prompt $p$, and for sets $A$ of levels and $P$ of prompts define
\[
S^\ast(A,P)\;=\;\bigcap_{a\in A,\;p\in P} S(a,p).
\]
Let $r_{a,p}(t)$ be mean success across the injected fault sites for task
$t$. We report three targets because they answer different questions.
The \emph{separately screened rate} averages $r_{a,p}(t)$ over $S(a,p)$.
It describes recovery on the work that one configuration handles
cleanly, but two such rates may use different tasks. The
\emph{matched-task recovery rate} averages over tasks in
$S(a_1,p)\cap S(a_2,p)$ when comparing precisions $a_1$ and $a_2$ under
prompt $p$. It isolates recovery on the same clean-solvable work. The
\emph{full-pipeline success rate} averages over all generated tasks and
assigns zero to a task that fails the clean screen. Operationally, it is
the success rate of a gated procedure in which a configuration is first
certified on each task without faults and then run on that task under
faults, with an uncertified task counted as a failure. This is the
protocol of fault-injection evaluations that only perturb tasks a
system already solves, including ours and ToolMaze's
\citep{zhu2026toolmaze}. Because an uncertified task might still
succeed under faults, the rate is a lower bound on ungated success, and
we use it only to rank configurations under the gated protocol.

Our primary conditional comparison matches 8-bit and 4-bit tasks within
each prompt, retaining 9--20 Llama tasks and 4--10 Qwen tasks. As a
stricter sensitivity analysis, we also use
$S^\ast(\{\texttt{Q8},\texttt{Q4}\},\mathcal{P})$, the tasks that pass at
both precisions under all five prompts. This intersection contains nine
Llama tasks and three Qwen tasks. The exploratory 3-bit analysis uses
separate Llama-only intersections (Appendix~\ref{app:promptinteraction}).

\section{Experimental setup}
\label{sec:method}

\paragraph{Tasks.} The main prompt experiments use twenty deterministic
tasks in a synthetic order-management environment. Ten tasks require
lookups of an order, customer, and discount, and ten add a shipping-fee
lookup.
Values come from a fixed seed, and the generator verifies each gold
answer and minimal trajectory. A broader forty-task suite with multi-order and distractor-tool task
families appears only in the appendix $P_0$ ladder.

\paragraph{Models and serving.} The main experiments evaluate
Llama-3.1-8B-Instruct \citep{grattafiori2024llama3} at \texttt{Q8\_0},
\texttt{Q4\_K\_M}, and \texttt{Q3\_K\_M}, and
Qwen2.5-7B-Instruct \citep{yang2024qwen25} at \texttt{Q8\_0} and
\texttt{Q4\_K\_M}. The
$P_0$-only appendix ladder also includes Llama \texttt{Q6\_K},
\texttt{Q5\_K\_M}, and \texttt{Q2\_K}. We serve all checkpoints with
llama.cpp \citep{llamacpp}, full GPU offload, temperature 0, top-$k$ 1,
and a fixed seed on one NVIDIA L4. Both result files record the pinned
llama.cpp commit (\texttt{b10f9ca}) and Hugging Face model-file revision.

\paragraph{Tool protocol and faults.} On each turn the model emits one
JSON object, the harness executes the requested tool, and the next prompt
includes the observation. An episode is one attempt at one task. It ends
at \texttt{submit\_answer},
\texttt{report\_failure}, or the step budget. Following ToolMaze
\citep{zhu2026toolmaze}, the transient-error condition (\qet) makes one
target call fail while a later identical call succeeds. We inject this
fault at every valid site on the gold tool path, producing 13--50 fault
episodes per prompt--precision cell. Displaced (\qit) and permanent (\qep)
faults were evaluated only under $P_0$ (Appendix~\ref{app:ladder}).

\paragraph{Output parsing and scoring.}
\label{par:scoring}
All prompts ask for one JSON object per turn and a literal numeric
amount in \texttt{submit\_answer}. The harness is tolerant on both
counts. It extracts the first JSON object embedded in a non-JSON turn,
and it evaluates a submitted arithmetic expression such as
\texttt{200.0 * 0.5} with a restricted parser. Each episode log records
whether the episode was strict JSON throughout and whether the answer
was an expression, so the released data can be rescored. We call the
main-results policy \emph{tolerant scoring} and the alternative that
requires strict JSON and a literal amount \emph{strict scoring}; the
clean screen uses the same executor, so strict scoring also changes
which tasks pass it. Section~\ref{sec:scoring} reports all three
targets under both policies.

\begin{figure}[t]
\centering
\includegraphics[width=0.8\linewidth]{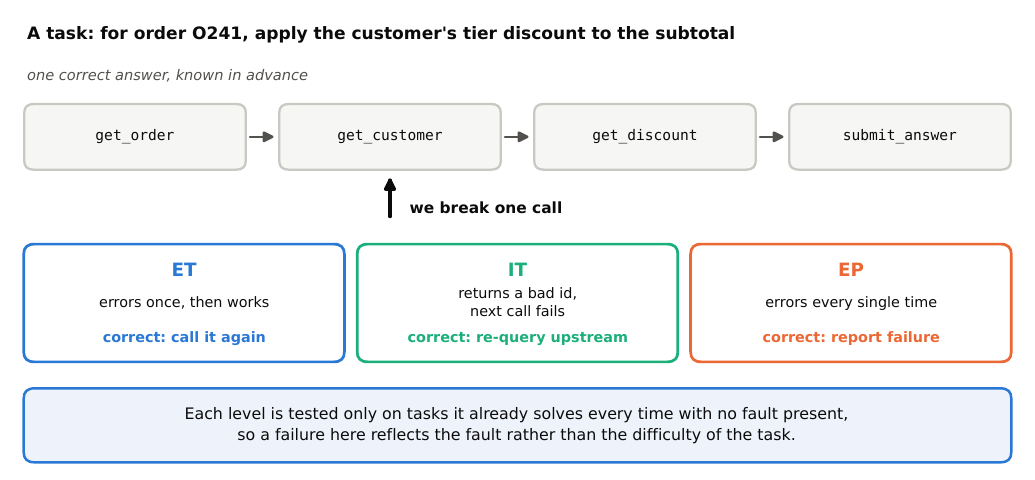}
\caption{\textbf{One task and the three fault conditions.} Each task is a
deterministic tool chain with one verified answer, and a fault is injected
at one call on the required path. The transient condition (\qet) is the
main experiment. Displaced (\qit) and permanent (\qep) faults are
evaluated under $P_0$ only (Appendix~\ref{app:ladder}). Fault episodes run
only on tasks that pass that condition's clean screen.}
\label{fig:task}
\end{figure}

\paragraph{Prompts.} $P_0$ contains tool signatures, a
one-object-per-turn format rule, a worked example, an instruction that
tools may fail and may be retried, and \texttt{report\_failure}. We
wrote four rewrites that retain the tool interface and output format
while changing the instructions. $P_1$ and $P_2$ use prose, $P_3$ uses
numbered rules, and $P_4$ uses terse imperatives. These rewrites
are not controlled paraphrases. They differ along several dimensions, and
only $P_0$ explicitly requires repeated attempts before a tool may be
declared permanently unavailable. Moreover, $P_3$ and $P_4$ were written
after the $P_1$/$P_2$ results were known. The supplementary material
contains the complete prompts.

\paragraph{Screening and execution order.} Each prompt--precision cell
receives one clean pass over all twenty tasks, and faults are injected only
into tasks that pass. After the exploratory Llama sweep produced the
screening pattern, we froze its prompts, code, and analysis plan before
running a pinned $5{\times}3$ sweep with complete episode logging. Its
clean-task sets and fault-success counts match the released exploratory
sweep in all fifteen cells. We then ran a $5{\times}2$ Qwen extension with
the same tasks, prompts, harness, decoding settings, screening rule, and
analysis. The Qwen extension was planned after the Llama result and is
therefore a cross-model extension rather than a preregistered replication.
Because decoding is deterministic, repeated identical counts check the
pipeline but do not provide independent statistical samples.

\paragraph{Uncertainty.} Rates carry episode-level 95\% Wilson intervals
where shown, while all main contrasts use task means. Comparisons between
different clean-task sets use a two-sample task bootstrap. Fixed-task comparisons use paired task bootstraps and two-sided sign
tests that exclude ties. Every bootstrap uses 20{,}000 samples. Tasks, rather than
injected fault sites or deterministic reruns, are the unit of uncertainty.
The Llama per-level $P_0$--$P_1$ comparisons in
Appendix~\ref{app:promptinteraction} were specified after inspection of the
pre-specified outcomes and are therefore exploratory.

\section{Results}
\label{sec:results}

\subsection{No precision level is consistently better}
\label{sec:matched}

On tasks that both precisions pass cleanly under the same prompt,
Llama's 8-bit--4-bit recovery difference is $+20.2$ points under $P_0$
(95\% CI $[+4.8,+39.3]$), $+3.9$ under $P_1$, $+4.4$ under $P_2$,
$+4.2$ under $P_3$, and $0$ under $P_4$
(Figure~\ref{fig:collapse}B). Only the $P_0$ interval excludes zero.
Thus the Llama estimates are compatible with an 8-bit advantage under
one prompt, but they do not show a stable difference across prompts.

Qwen varies in both direction and magnitude on the prompt-matched tasks.
Its 8-bit--4-bit differences are $+8.3$, $-3.3$, $-10.0$, $-50.0$, and
$+35.0$ points from $P_0$ through $P_4$. The intervals exclude zero under
$P_3$ (CI $[-65.0,-35.0]$) and $P_4$ (CI $[+20.0,+50.0]$), which favor
opposite precisions. This result rules out a single 8-bit or 4-bit
recovery advantage that generalizes across the two tested models and
five prompts.

Full-pipeline success includes ordinary task competence as well as
fault recovery (Figure~\ref{fig:collapse}C and
Table~\ref{tab:endtoend}). For Llama, the 8-bit point estimate is higher
under all five prompts; the intervals exclude zero under $P_0$
($+27.5$ points, CI $[+12.5,+44.2]$) and $P_4$ ($+28.3$, CI
$[+6.7,+48.3]$). Qwen again changes direction, with differences of
$+33.3$, $-1.7$, $-8.3$, $-28.3$, and $+17.5$ points. The $P_0$, $P_3$,
and $P_4$ intervals exclude zero. A deployment decision based on this
pipeline would therefore depend on both the model and the prompt.

The stricter all-prompt intersection gives the same qualitative warning
with less power. It retains nine Llama tasks and only three Qwen tasks;
all intervals include zero, and the selected tasks often sit near the
recovery ceiling. We use it as a sensitivity analysis rather than as the
primary estimate.

\subsection{Separate clean screening can change the conclusion}
\label{sec:ownset}

Table~\ref{tab:sweep} reports recovery when each configuration is scored
on its own clean-passing tasks. For Llama, the task-mean 8-bit--4-bit
difference is $+14.4$ points under $P_0$, $+0.6$ under $P_1$, $-0.7$
under $P_2$, $+4.2$ under $P_3$, and $-17.5$ under $P_4$. Only the $P_4$
two-sample task-bootstrap interval excludes zero (95\% CI
$[-29.8,-7.0]$). Under that prompt, however, 8-bit passes 19 tasks
cleanly and 4-bit passes 10, so the two rates summarize different work.

Qwen's separately screened differences are $+17.0$, $-3.3$, $-7.6$,
$-47.6$, and $+35.0$ points from $P_0$ through $P_4$. The intervals
exclude zero under $P_0$ (CI $[+0.6,+33.3]$), $P_3$ (CI
$[-62.0,-31.8]$), and $P_4$ (CI $[+10.0,+55.0]$). The clean sets differ
under $P_0$ (11 tasks for 8-bit and 5 for 4-bit) but coincide under
$P_1$ and $P_4$. Consequently, some separately screened comparisons mix
task selection with recovery, while others already compare the same
tasks.

\begin{table}[t]
\centering
\caption{Task-mean transient-fault recovery on each condition's own
clean-solvable set. Brackets give pooled successful/total fault episodes.
Rows must not be compared as recovery effects unless they use the same
tasks.}
\label{tab:sweep}
\footnotesize
\begin{tabular}{llccccc}
\toprule
Model & Level & $P_0$ & $P_1$ & $P_2$ & $P_3$ & $P_4$\\
\midrule
Llama & \texttt{Q8\_0}    & 87.0\% [37/44] & 89.5\% [41/47] & 72.2\% [29/44] & 94.2\% [47/50] & 82.5\% [38/48]\\
      & \texttt{Q4\_K\_M} & 72.6\% [21/32] & 88.9\% [38/44] & 72.9\% [25/38] & 90.0\% [44/50] & 100\% [20/20]\\
      & \texttt{Q3\_K\_M} & 41.2\% [16/47] & 0\% [0/32] & 76.2\% [22/32] & 61.1\% [26/44] & 85.0\% [17/21]\\
\addlinespace
Qwen  & \texttt{Q8\_0}    & 97.0\% [23/24] & 90.0\% [18/21] & 83.3\% [17/21] & 43.3\% [9/21]  & 85.0\% [17/21]\\
      & \texttt{Q4\_K\_M} & 80.0\% [10/13] & 93.3\% [19/21] & 90.9\% [21/24] & 90.9\% [21/24] & 50.0\% [10/21]\\
\bottomrule
\end{tabular}
\end{table}

Llama/$P_4$ demonstrates the largest change in interpretation. Separate
screening favors 4-bit by $17.5$ points; the nine prompt-matched tasks
give exactly zero; and the full pipeline favors 8-bit by $28.3$ points.
The first comparison is therefore driven by which tasks enter each
denominator rather than by a 4-bit recovery advantage on shared work.
Llama/$P_0$ shows the opposite failure mode, where matching fourteen
tasks reveals a $+20.2$-point difference that the noisier separately screened
estimate does not resolve.

Qwen shows both selected-task and persistent recovery effects. Under
$P_0$, the estimate falls from $+17.0$ points on separate task sets to
$+8.3$ on the four prompt-matched tasks, while full-pipeline success is
$+33.3$ points. Under $P_3$ the 8-bit clean set (ten tasks) is a subset
of the 4-bit set (eleven tasks), and under $P_4$ the two sets coincide;
in both cases the matched population is the same ten tasks, and the
large, oppositely signed differences ($-50.0$ and $+35.0$ points)
survive matching. Separate screening can thus
create, hide, or exaggerate a comparison, but it does not explain every
precision difference in these data.

\subsection{Scoring policy changes the Llama comparison}
\label{sec:scoring}

The results above use tolerant scoring (Section~\ref{sec:method}).
Table~\ref{tab:strict} rescores the same logs under strict scoring,
with the clean screen rescored under the same policy; no model was
rerun. Qwen almost never submits an expression (one of 165 successful
fault episodes), so its estimates are essentially unchanged and the
$P_3$/$P_4$ reversal is identical under both policies. Llama submits expressions heavily under every prompt, and under
$P_4$ the two precisions differ sharply. There, 30 of the 38 successful
8-bit fault episodes submit an expression, against 3 of 20 for 4-bit. Strict scoring therefore removes most of 8-bit Llama's
successes, shrinks its clean sets from 18--20 tasks to 4--10, and
reverses the $P_4$ comparison on every target. Separate screening
moves from $-17.5$ to $-42.9$ points, the matched-task difference from
$0$ to $-33.3$ (six tasks), and full-pipeline success from $+28.3$ to
$-15.0$. The Llama full-pipeline point estimate no longer favors 8-bit
under every prompt ($+12.5$, $+10.0$, $0$, $0$, $-15.0$ from $P_0$ to
$P_4$).

\begin{table}[t]
\centering
\caption{8-bit minus 4-bit difference in percentage points under
tolerant scoring (main results) and strict scoring (strict JSON, literal
amount, clean screen rescored), for the separately screened,
matched-task ($n$ matched tasks), and full-pipeline targets. Brackets
are 95\% task-bootstrap intervals under strict scoring; tolerant
intervals appear in Figure~\ref{fig:collapse} and the text.}
\label{tab:strict}
\footnotesize
\setlength{\tabcolsep}{4pt}
\begin{tabular}{llrrrlrl}
\toprule
 & & \multicolumn{2}{c}{Separate} & \multicolumn{2}{c}{Matched ($n$)} & \multicolumn{2}{c}{Full pipeline}\\
\cmidrule(lr){3-4}\cmidrule(lr){5-6}\cmidrule(lr){7-8}
Model & Prompt & tol. & strict & \multicolumn{1}{c}{tol.} & \multicolumn{1}{c}{strict} & \multicolumn{1}{c}{tol.} & \multicolumn{1}{c}{strict}\\
\midrule
Llama & $P_0$ & $+14.4$ & $+54.2$ & $+20.2$ (14) & $+50.0$ (3) $[0,+100]$     & $+27.5$ & $+12.5$ $[0,+27.5]$\\
      & $P_1$ & $+0.6$  & $+12.8$ & $+3.9$ (17)  & $+22.2$ (9) $[0,+44.4]$    & $+5.0$  & $+10.0$ $[0,+22.5]$\\
      & $P_2$ & $-0.7$  & $0$     & $+4.4$ (15)  & $0$ (10) $[0,0]$           & $+6.7$  & $0$ $[0,0]$\\
      & $P_3$ & $+4.2$  & $-15.0$ & $+4.2$ (20)  & $-25.0$ (4) $[-50.0,0]$    & $+4.2$  & $0$ $[-10.0,+12.5]$\\
      & $P_4$ & $-17.5$ & $-42.9$ & $0$ (9)      & $-33.3$ (6) $[-66.7,-8.3]$ & $+28.3$ & $-15.0$ $[-30.0,-2.5]$\\
\addlinespace
Qwen  & $P_0$ & $+17.0$ & $+13.9$ & $+8.3$ (4)   & $0$ (4) $[0,0]$            & $+33.3$ & $+31.7$ $[+10.0,+55.0]$\\
      & $P_1$ & $-3.3$  & $-3.3$  & $-3.3$ (10)  & $-3.3$ (10) $[-10.0,0]$    & $-1.7$  & $-1.7$ $[-5.0,0]$\\
      & $P_2$ & $-7.6$  & $-7.6$  & $-10.0$ (10) & $-10.0$ (10) $[-25.0,0]$   & $-8.3$  & $-8.3$ $[-18.3,0]$\\
      & $P_3$ & $-47.6$ & $-47.6$ & $-50.0$ (10) & $-50.0$ (10) $[-65.0,-35.0]$ & $-28.3$ & $-28.3$ $[-42.5,-15.0]$\\
      & $P_4$ & $+35.0$ & $+35.0$ & $+35.0$ (10) & $+35.0$ (10) $[+20.0,+50.0]$ & $+17.5$ & $+17.5$ $[+7.5,+27.5]$\\
\bottomrule
\end{tabular}
\end{table}

Executor leniency is thus a third design choice, alongside the prompt
and the evaluation target, that can reverse a quantization comparison,
and under strict scoring Llama/$P_4$ favors 4-bit on all three targets.
The two policies also measure different things. Tolerant scoring asks
whether the agent recovers the right number; strict scoring also asks
whether it obeys the output contract, which matters when a downstream
parser does no repair. Because 8-bit Llama violates the literal-amount instruction far more
often than 4-bit Llama under $P_4$, a strict executor penalizes it
there for format compliance rather than recovery. We keep
tolerant scoring as the main result because it isolates recovery, but
a deployment comparison should use the deployed executor's policy and
say so.

\section{Implications for evaluation}
\label{sec:implications}

\paragraph{Match the measure to the claim.} A conditional statement about
recovery requires the same tasks in every compared configuration. A
prompt-wise clean-task intersection provides that comparison, although
the intersection may be small or selected for easier tasks. A deployment
comparison should additionally report full-pipeline success over the
entire suite, because clean competence is part of system performance.

\paragraph{Use tasks as the unit of evidence.} Multiple injected sites
within one task share the task's structure and should not be treated as
independent evidence about a cross-condition difference. Deterministic reruns check reproducibility rather than increasing sample
size, and reporting task-level intervals alongside episode counts makes
these dependencies visible.

\paragraph{Evaluate the deployed instructions.} Prompts that change when
an agent should retry, stop, or declare failure specify different
policies rather than interchangeable wording. Recovery should therefore
be tested with the deployment prompt at each candidate precision.
Attributing a difference to one instruction requires a single-factor
ablation rather than a comparison of multi-feature rewrites.

\paragraph{State the scoring policy.} Output repair in the harness is
part of the evaluated system. It should be documented, logged per
episode so that results can be rescored, and matched to the deployed
executor. When variants differ in how often they rely on repair, as 8-bit and 4-bit Llama do under $P_4$, the policy can decide the
comparison,
and reporting both makes that dependence visible.

\section{Limitations}
\label{sec:limits}

Twenty tasks in one synthetic environment and two models of similar size
are insufficient to estimate how often these patterns occur in other
domains, model families, or scales. The twenty tasks come from two
closely related templates; the generator's multi-order and
distractor-tool families were run only in the $P_0$ ladder, so the key
comparisons are untested on other task structures. The all-prompt
8-bit--4-bit intersections contain nine Llama tasks and three Qwen
tasks and show substantial ceiling effects. They cannot establish
equivalence or characterize the full clean-passing population. The
fault model is also narrow, consisting of one transient failure that
an identical retry resolves, without repeated failures,
backoff, rate limits, or tool state that changes between attempts.

The five author-written prompts are not controlled paraphrases. $P_3$
and $P_4$ were composed after early results were known, and the per-level
$P_0$--$P_1$ analysis was specified after the pre-specified outcomes.
Because $P_0$ and $P_1$ differ in more than the retry criterion, the
observed interaction does not establish a causal mechanism.

Finally, \texttt{Q8\_0}, \texttt{Q4\_K\_M}, and \texttt{Q3\_K\_M} are
GGUF presets that differ in quantizer construction as well as nominal
bit budget; ``8-bit'' and ``4-bit'' in this paper are shorthand for
these presets, and our conclusions apply to them rather than to bit
width in isolation or to other quantization families. Qwen was
evaluated only at \texttt{Q8\_0} and \texttt{Q4\_K\_M}, so the
exploratory 3-bit prompt failure has not been tested in a second model.
The strict-scoring analysis rescored existing logs without rerunning
inference; a model told that its expression was rejected might behave
differently. Larger task suites, independently designed prompts,
additional model families and quantization methods, and the
single-factor retry-policy ablation are needed to determine how
broadly the observed variation generalizes.

\section{Conclusion}

We find no consistent \texttt{Q8\_0} or \texttt{Q4\_K\_M} recovery
advantage across Llama, Qwen, and five prompts. Results vary with the
model, the prompt, whether the target is matched-task recovery or
full-pipeline success, and whether the executor repairs
non-conforming output. Separate clean screening can change the
conclusion by changing the evaluated tasks, and strict scoring can
reverse it by removing successes that depend on repair, as Llama/$P_4$
demonstrates on both counts. Evaluations should report both targets
across prompts, state and log the scoring policy, and quantify
uncertainty across tasks.

\bibliographystyle{plainnat}
\bibliography{refs}

\clearpage
\appendix

\section{Full Llama quantization ladder under the original prompt}
\label{app:ladder}

\begin{figure}[h]
\centering
\begin{tikzpicture}
\begin{axis}[
  width=\linewidth,
  height=5.4cm,
  ymin=0,
  ymax=1.08,
  ylabel={Success rate},
  xlabel={GGUF quantization level},
  symbolic x coords={Q8,Q6,Q5,Q4,Q3},
  xtick=data,
  xticklabels={\texttt{Q8\_0},\texttt{Q6\_K},\texttt{Q5\_K\_M},\texttt{Q4\_K\_M},\texttt{Q3\_K\_M}},
  tick label style={font=\scriptsize},
  label style={font=\small},
  ymajorgrids=true,
  grid style={gray!25},
  legend style={at={(0.5,1.19)},anchor=south,legend columns=2,draw=none,font=\scriptsize},
  error bars/y dir=both,
  error bars/y explicit,
]
\addplot+[gray!70!black,mark=o,thick]
coordinates {(Q8,1.000) += (0,0.000) -= (0,0.168)
             (Q6,1.000) += (0,0.000) -= (0,0.168)
             (Q5,1.000) += (0,0.000) -= (0,0.168)
             (Q4,1.000) += (0,0.000) -= (0,0.215)
             (Q3,1.000) += (0,0.000) -= (0,0.194)};
\addlegendentry{Clean (clean-solvable set)}
\addplot+[blue!70!black,mark=o,thick]
coordinates {(Q8,0.830) += (0,0.081) -= (0,0.131)
             (Q6,0.809) += (0,0.087) -= (0,0.134)
             (Q5,0.851) += (0,0.075) -= (0,0.128)
             (Q4,0.594) += (0,0.151) -= (0,0.171)
             (Q3,0.526) += (0,0.149) -= (0,0.154)};
\addlegendentry{Transient fault (\qet)}
\addplot+[green!60!black,mark=square*,thick]
coordinates {(Q8,0.763) += (0,0.107) -= (0,0.155)
             (Q6,0.684) += (0,0.125) -= (0,0.159)
             (Q5,0.684) += (0,0.125) -= (0,0.159)
             (Q4,0.714) += (0,0.133) -= (0,0.185)
             (Q3,0.781) += (0,0.109) -= (0,0.169)};
\addlegendentry{Corrupted data (\qit)}
\addplot+[orange!85!black,mark=triangle*,thick]
coordinates {(Q8,0.532) += (0,0.135) -= (0,0.140)
             (Q6,0.489) += (0,0.138) -= (0,0.137)
             (Q5,0.681) += (0,0.115) -= (0,0.143)
             (Q4,0.656) += (0,0.140) -= (0,0.173)
             (Q3,0.895) += (0,0.064) -= (0,0.136)};
\addlegendentry{Dead tool: abstain (\qep)}
\end{axis}
\end{tikzpicture}
\caption{The $P_0$ ladder: success by quantization level under clean
execution and the three fault conditions, each level on its own
same-prompt clean-solvable set, with episode-level 95\% Wilson intervals.
The \texttt{Q5\_K\_M}-to-\texttt{Q4\_K\_M} transient-recovery decrease is
conditional on $P_0$; compare Figure~\ref{fig:collapse}.}
\label{fig:ladder}
\end{figure}
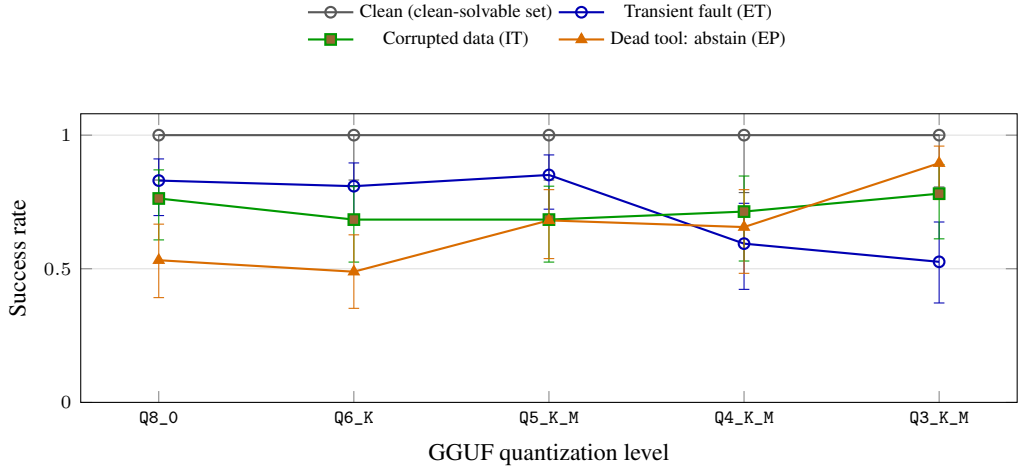

Before the prompt sweep, we evaluated six quantization presets under $P_0$
and three fault conditions (Figure~\ref{fig:ladder},
Table~\ref{tab:main}). For transient faults, pooled success on each
preset's $P_0$ clean-solvable set is $117/141=83\%$ for
\texttt{Q8\_0}--\texttt{Q5\_K\_M} and $39/70=56\%$ for
\texttt{Q4\_K\_M}--\texttt{Q3\_K\_M} (Fisher exact
$p=4.7{\times}10^{-5}$). On eleven tasks clean-solvable by both groups,
five favor the higher-precision group, six tie, and none favor the lower
group (sign test $p=0.063$, task-bootstrap CI $[5,29]$ points). Thus the
site-level evidence is stronger than the task-level evidence.

No clean-solvable-set permanent-fault episode above 3 bits produced a fabricated
numeric answer. Failures instead involved insufficient retries or
unstructured termination. The earlier one-prompt Qwen2.5-7B-Instruct check
gave 96\% versus 77\% pooled recovery. Its complete five-prompt extension
is now part of the main analysis, and the earlier file is retained only as
an audit trail. \texttt{Q2\_K} fails the pre-specified minimum of
eight clean tasks and is excluded. It nevertheless emits strictly
well-formed tool calls in 102 of 120 clean episodes (a tolerant parser
recovers a call in all 120), usually repeating one call until the step
budget ends. In an earlier protocol iteration, adding one sentence about
tool resilience reduced the 4-bit model's clean accuracy by 26 points while
leaving the 8-bit result unchanged. Because prompt length and termination
signaling both changed, we do not attribute that observation to one cause.

\begin{table}[h]
\centering
\caption{Results under $P_0$ on each level's same-prompt clean-solvable
set. \qep\ success denotes correct \texttt{report\_failure} use;
``Invented'' counts fabricated numeric answers in \qep.}
\label{tab:main}
\small
\begin{tabular}{lcccccc}
\toprule
Level & CLEAN & \qet & \qit & \qep & Invented\\
\midrule
\texttt{Q8\_0}   & 19/19 (100\%) & 39/47 (83\%) & 29/38 (76\%) & 25/47 (53\%) & 0\\
\texttt{Q6\_K}   & 19/19 (100\%) & 38/47 (81\%) & 26/38 (68\%) & 23/47 (49\%) & 0\\
\texttt{Q5\_K\_M}& 19/19 (100\%) & 40/47 (85\%) & 26/38 (68\%) & 32/47 (68\%) & 0\\
\texttt{Q4\_K\_M}& 14/14 (100\%) & 19/32 (59\%) & 20/28 (71\%) & 21/32 (66\%) & 0\\
\texttt{Q3\_K\_M}& 16/16 (100\%) & 20/38 (53\%) & 25/32 (78\%) & 34/38 (89\%) & 3\\
\bottomrule
\end{tabular}
\end{table}
\FloatBarrier

\section{Pipeline yield tables}
\label{app:yield}

\begin{table}[h]
\centering
\caption{Full-suite pipeline yield: task-mean transient-fault success over
all twenty tasks, with a task that fails the clean screen assigned zero.
This quantity combines clean competence and fault recovery.}
\label{tab:endtoend}
\small
\begin{tabular}{llccccc}
\toprule
Model & Level & $P_0$ & $P_1$ & $P_2$ & $P_3$ & $P_4$\\
\midrule
Llama & \texttt{Q8\_0}    & .78 & .85 & .65 & .94 & .78\\
      & \texttt{Q4\_K\_M} & .51 & .80 & .58 & .90 & .50\\
      & \texttt{Q3\_K\_M} & .39 & .00 & .53 & .55 & .42\\
\addlinespace
Qwen  & \texttt{Q8\_0}    & .53 & .45 & .42 & .22 & .425\\
      & \texttt{Q4\_K\_M} & .20 & .47 & .50 & .50 & .25\\
\bottomrule
\end{tabular}
\end{table}

\FloatBarrier

\section{Exploratory Llama prompt--precision interaction}
\label{app:promptinteraction}

\begin{figure}[h]
\centering
\includegraphics[width=\linewidth]{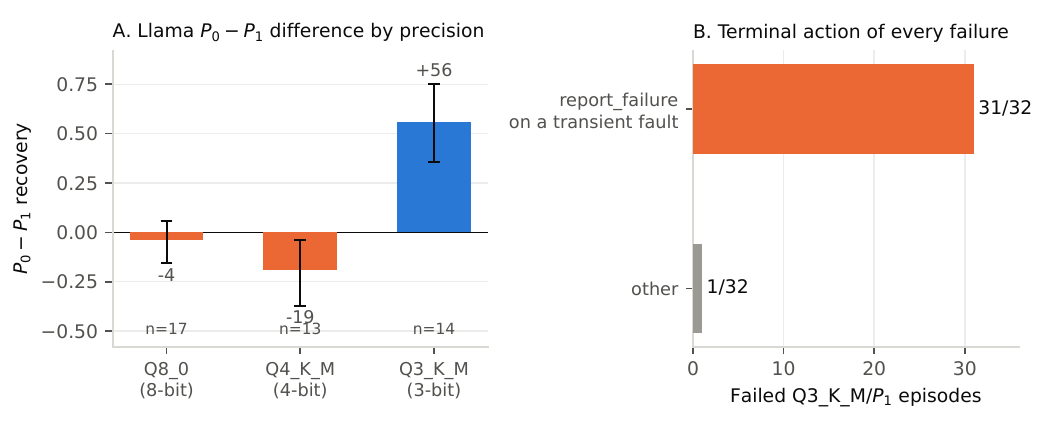}
\caption{\textbf{Exploratory Llama-only prompt--precision result.}
\textbf{A.} Task-mean $P_0$--$P_1$ recovery difference on each precision
level's pairwise common clean set, with task-bootstrap 95\% intervals.
\textbf{B.} Terminal actions of all 32 failed
\texttt{Q3\_K\_M}/$P_1$ episodes: 31 declare the transient fault permanent.}
\label{fig:policy}
\end{figure}

This analysis was specified after inspection of the planned Llama
outcomes, and $P_0$ and $P_1$ differ in several ways. On each precision
level's pairwise common clean set, the $P_0$--$P_1$ recovery difference is
$-3.9$ points at 8 bits ($n{=}17$ tasks, CI $[-15.7,+5.9]$), $-19.2$ at
4 bits ($n{=}13$, CI $[-37.2,-3.8]$, sign test $p=0.125$ with four untied
tasks), and $+56.0$ at 3 bits ($n{=}14$, CI $[+35.7,+75.0]$;
Figure~\ref{fig:policy}A). Thus $P_1$ has the higher point estimate at 8
and 4 bits, but the direction reverses at 3 bits.

On the distinct nine-task common set that passes every Llama 3-bit
prompt condition, $P_0$ recovers 77.8\% and $P_1$ recovers 0\%; all nine
tasks favor $P_0$ (sign test $p=3.9{\times}10^{-3}$, CI
$[+61.1,+94.4]$). The other prompts recover 100\% for $P_2$, 66.7\% for
$P_3$, and 94.4\% for $P_4$ on this set.

Thirty-one of the 32 failed 3-bit/$P_1$ episodes terminate with
\texttt{report\_failure} after a transient error; the remaining episode
submits an incorrect answer (Figure~\ref{fig:policy}B). This regularity
describes the output behavior, not an internal mechanism. A causal test
would change only the repeated-attempts criterion between $P_0$ and $P_1$
and rerun the comparison at all three precision levels. Qwen was not run at
3 bits, and this ablation has not been conducted.

\FloatBarrier

\clearpage
\section*{NeurIPS Paper Checklist}
\begin{enumerate}\itemsep2pt\small
\item \textbf{Claims.} Yes. The abstract and introduction state one
primary empirical claim: neither 8-bit nor 4-bit quantization has a
consistent tool-failure recovery advantage across the tested models and
prompts. The paper also shows that separate clean screening can change
the interpretation by changing the evaluated tasks. The Llama
prompt--precision interaction is labeled exploratory, and no causal claim
is made about which prompt feature drives it.
\item \textbf{Limitations.} Yes. Section~\ref{sec:limits} covers the small intersections,
one synthetic environment, two similarly sized models, author-written
prompts, post-hoc $P_3$/$P_4$, ceiling effects, the un-run criterion
ablation, GGUF preset coverage, and the post-Llama timing of the Qwen
extension.
\item \textbf{Theory assumptions and proofs.} N/A. No theoretical results.
\item \textbf{Experimental reproducibility.} Yes. Fixed seeds,
temperature-0 decoding, pinned llama.cpp commit and model-file revision
recorded in both released result files, and a Llama analysis plan fixed
before its pinned run. The Qwen extension uses the frozen protocol but was
planned after the Llama result and is not described as preregistered.
\item \textbf{Open access to data and code.} Yes. The task generator,
harness, prompts, analysis code, and complete per-episode logs for all
twenty-five Llama and Qwen sweep cells are in the supplementary ZIP;
\texttt{confirmatory\_paper\_stats.py} reproduces the reported
cross-prompt comparisons from the shipped data, and
\texttt{strict\_vs\_tolerant.py} reproduces Table~\ref{tab:strict}.
\item \textbf{Experimental settings.} Yes. Section~4 and the supplementary
README give models, quantization levels, decoding parameters, prompts, and
hardware.
\item \textbf{Error bars.} Yes. Episode-level 95\% Wilson intervals where
shown; two-sample task-bootstrap intervals for contrasts between different
clean-task sets; and paired task-bootstrap intervals with sign tests for
fixed-task contrasts. Every bootstrap uses 20{,}000 samples. Exploratory
analyses are labeled, and the text notes where the bootstrap and sign test
disagree.
\item \textbf{Compute.} Yes. All experiments ran on a single NVIDIA L4
(24\,GB), roughly 12 GPU-hours total across the exploratory runs, pinned
Llama run, and Qwen extension.
\item \textbf{Code of ethics.} Yes. The research conforms to the NeurIPS
Code of Ethics.
\item \textbf{Broader impacts.} The work aims to prevent misleading
reliability conclusions when quantized agents are selected for deployment.
We foresee no negative applications specific to this evaluation method.
\item \textbf{Safeguards.} N/A. No models or data with misuse risk are
released.
\item \textbf{Licenses.} Yes. Llama-3.1 (Llama 3.1 Community License),
Qwen2.5 (Apache 2.0), llama.cpp (MIT), community GGUF conversions credited.
\item \textbf{New assets.} Yes. The released task suite and harness are
documented in the supplementary README.
\item \textbf{Crowdsourcing / human subjects.} N/A.
\item \textbf{IRB.} N/A.
\item \textbf{LLM usage.} LLM assistance was used for experiment code,
analysis tooling, and draft editing. All experimental design decisions,
results, and citations were verified by the author, who takes full
responsibility for the content. No LLM is an author.
\end{enumerate}

\end{document}